\documentclass[conference]{IEEEtran}
\IEEEoverridecommandlockouts
\usepackage{cite}
\usepackage{amsmath,amssymb,amsfonts}
\usepackage{algorithmic}
\usepackage[]{algorithm2e}
\usepackage{graphicx}
\usepackage{textcomp}
\usepackage{xcolor}
\usepackage[switch]{lineno}
\usepackage{tikz}

\newcommand\copyrighttext{%
  \footnotesize \textcopyright 2021 IEEE. Personal use of this material is permitted.
  Permission from IEEE must be obtained for all other uses, in any current or future
  media, including reprinting/republishing this material for advertising or promotional
  purposes, creating new collective works, for resale or redistribution to servers or
  lists, or reuse of any copyrighted component of this work in other works.
  DOI: not yet}
\newcommand\copyrightnotice{%
\begin{tikzpicture}[remember picture,overlay]
\node[anchor=south,yshift=10pt] at (current page.south) {\fbox{\parbox{\dimexpr\textwidth-\fboxsep-\fboxrule\relax}{\copyrighttext}}};
\end{tikzpicture}%
}

\def\BibTeX{{\rm B\kern-.05em{\sc i\kern-.025em b}\kern-.08em
    T\kern-.1667em\lower.7ex\hbox{E}\kern-.125emX}}

\begin{document}

\title{Multi-objective discovery of PDE systems using evolutionary approach\\
}

\author{\IEEEauthorblockN{Mikhail Maslyaev}
\IEEEauthorblockA{\textit{NSS (Nature systems simulation) Lab} \\
\textit{ITMO University}\\
Saint-Petersburg, Russia \\
mikemaslyaev@itmo.ru}
\and
\IEEEauthorblockN{Alexander Hvatov}
\IEEEauthorblockA{\textit{NSS Lab} \\
\textit{ITMO University}\\
Saint-Petersburg, Russia \\
alex\_hvatov@itmo.ru}
}

\maketitle
\copyrightnotice

\begin{abstract}
Usually, the systems of partial differential equations (PDEs) are discovered from observational data in the single vector equation form. However, this approach restricts the application to the real cases, where, for example, the form of the external forcing is of interest. In the paper, a multi-objective co-evolution algorithm is described.  The single equations within the system and the system itself are evolved simultaneously to obtain the system. This approach allows discovering the systems with the form-independent equations. In contrast to the single vector equation, a component-wise system is more suitable for expert interpretation and, therefore, for applications. The example of the two-dimensional Navier-Stokes equation is considered. 
\end{abstract}


\section{Introduction}


Data-driven partial differential equation (PDE) discovery may link classical methods and modern machine learning methods. If the data-driven model is expressed in the form of differential equations, on the one hand, the classical analysis methods are applied. On the other hand, such a form is understandable by scientists in various application fields.

Single PDE discovery algorithms are usually applied sparse regression to the pre-defined library of terms. We proposed the evolutionary-based algorithm that allows combining evolutionary optimization and sparse regression to solve symbolic regression problems effectively.

Transfer to the multi-objective optimization makes the resulting model form more flexible. For example, the model's complexity may be controlled, and the Pareto frontier of the model with different complexity and data reproduction levels may be obtained. In a more general multi-objective discovery algorithm case, the partial differential equations systems may be obtained.

Modern equation discovery methods usually treat systems as a single equation in a vector form \cite{b6}. However, this restricts the application of the system discovery. Systems overall are not interesting since their amount used in modern physics is established and restricted. It is often of interest to obtain the forcing function or additional parametrizing equation, which is impossible in the vector form case. Regression methods are hardly applicable in non-vectorizable system discovery. Evolutionary multi-objective system discovery may give space for various applications since the system equations are discovered separately.

The paper describes the combined method that uses multi-objective co-evolutionary optimization (simultaneous equations and the system evolution) and the sparse regression to obtain the partial differential equations system.

The paper is structured as follows: Sec.~\ref{sec:literature_rewiev} contains a brief review of the equation and their systems discovery algorithms; Sec.~\ref{sec:problem_statement} is dedicated to the mathematical description of the multi-objective system discovery; Sec.~\ref{sec:algorithm_description} describes the particular realization of the algorithm; Sec.~\ref{sec:experiment} briefly highlights the results obtained on a synthetic data - two-dimensional Navier-Stokes equation solution; Sec.~\ref{sec:conclusion} outlines the paper and also sketches the future work directions. All required data and code that are required to reproduce the results are available from the URL placed in Sec.~\ref{sec:repo}

\section{Related work}
\label{sec:literature_rewiev}

Single differential equation discovery is of interest of many references \cite{b1,b2,b3,b4}. There are two main directions of differential equation discovery: sparse regression \cite{b1,b2} and neural networks \cite{b3,b4}.

Sparse regression consists of differential terms library definition and following sparse regression on the defined set. This approach has a speed drawback since the terms library should be as extensive as possible to cover all the possible differential equation types. The larger the set is, the slower is optimization process in sparse regression. However, this approach is well-developed and has different application areas.

One of the main advantages of the sparse regression methods is the straightforward mechanism of the equation discovery. In particular, by changing the sparsity hyperparameter, one may obtain the trade-off between complexity vs. quality equation.

Neural networks are fast and popular since there are a lot of established frameworks for neural network learning that allows the scientist to build the neural network without in-depth knowledge in the field \cite{keras}. However, they are susceptible to the training data. Thus, an additional filter network is required for the PDE discovery algorithm to be functional \cite{b3}. As a drawback, the results cannot be tuned to obtain the model with desired parameters. Also, the way to obtain the result is hardly interpretable, and the obtained equation is formed using ``blackbox'' in the form of the neural network.

We proposed the algorithm for the single equation discovery \cite{b5} that combines evolutionary optimization to obtain new terms for the sub-libraries that define the form of the equation and the sparse regression to filter out the insignificant terms in the sub-library to keep it as small as possible. Such an approach may be considered a symbolic regression version with additional evolutionary operators to obtain new terms.

The PDE systems are usually discovered as the equations in the vector form \cite{b6} meaning the methods for the single equation are applied to the vector values. The vector form restricts the type and the form of the obtained equation. Nevertheless, the ODE field situation is more straightforward, and we know the solution to obtain ``pure'' systems of the differential equations.

It should be emphasized that the problem of ordinary differential equation (ODE) systems in ``pure'' form solution is partially shown in \cite{b7}. However, due to the non-Markovian models' specificity, this approach cannot be applied to the general ODE and PDE systems discovery. 

\section{Problem statement}
\label{sec:problem_statement}

The standard problem formulation of the data-driven equation discovery assumes that equation is restored from the discrete datafield $D=\{(u_1(\vec{x}_i),...,u_n(\vec{x}_i)),i=1,2,...,M\}$, where $M$ is the number of the observation datapoints $\vec{x}_i \in \Omega \subset R^m$ avaliable. It is assumed that $D$ is the discrete analogue of the continuous field $\vec{u}(\vec{x})=(u_1(\vec{x}),...,u_n(\vec{x})) \,, \vec{u}: R^m \to R^n $ that represent the observation recordings of a physical process. Resulting discovered equation's system of $k$ equations has the form:

\begin{equation}
    S(\vec{u})=\left\{\begin{array}{cc}
         L_1(\vec{u})=0  \\
         ... \\
         L_k(\vec{u})=0
    \end{array}
    \right.
\label{eq:cont_system}
\end{equation}
  
In Eq.~\ref{eq:cont_system} single differential operator $L_i \in Eq$ represents the single differential equation, $Eq$ is the set of all possible equations that could be obtained with the given algorithm. Since Eq.~\ref{eq:cont_system} is assumed to be the system, all equation are assumed to be fullfiled simultaneously. In general, equation system $\bar{S}$ discovery task in an optimization problem formulated as:

\begin{equation}
\label{eq:theoretical_formulation}
    \bar{S}=\text{arg} \min \limits_{S \in Eq^k} S(\vec{u})
\end{equation}

In Eq.~\ref{eq:theoretical_formulation} $Eq^k=Eq \times ...\times Eq$ is the cartesian product of the sets of the possible equations. We emphasize that since only the discrete number of the points given, the operators are replaced with the discrete analogs such as finite differences. And the minimization task is reformulated as

\begin{equation}
    \forall i \, \bar{S}=\text{arg} \min \limits_{S \in Eq^k} S(\vec{u}_i) \, ,\vec{u}_i \in D
\label{eq:min_global}
\end{equation}

In practice, such formulation (Eq.~\ref{eq:min_global}) is hard to apply to the given method. Therefore, it is often rewritten as

\begin{equation}
     \bar{S}=\text{arg} \min \limits_{S \in Eq^k} \sum \limits_{i=1}^{i=M} || S(\vec{u}_i) ||
\label{eq:min_practical}
\end{equation}

In Eq.~\ref{eq:min_practical} norm $||\cdot||$ is chosen concerning problem specifics.

Introducing the multi-objective formulation allows tuning the discovered system in various ways. For example, for some problems, the data reproduction precision is less critical than the equation complexity.

The first group of the objectives we refer to as ``quality''. For a given equation $L$, the quality metric is the data reproduction norm that is represented as

\begin{equation}
     Q(L_j)= \sum \limits_{i=1}^{i=M} || L_j(\vec{u}_i) ||
\label{eq:norm_quality}
\end{equation}

The second group of objectives we refer to as ``complexity''. For a given equation $L$, the complexity metric is bound to the number of the differential terms in the equation that is denoted as $\#(L)$

\begin{equation}
     C(L_j)= \#(L_j)
\label{eq:norm_complexity}
\end{equation}

Two functions for every of the $k$ equations in the system form the space of the multi-objective optimization objectives. The next step is to define multi-objective optimization operator Eq.~\ref{eq:multiobj_op} and solve multi-objective optimization problem.

\begin{multline}
    F(L_1,...,L_k)=G(Q(L_1),C(L_1),...,Q(L_k),C(L_k)), \\
    F: R^{2k}->R^l
    \label{eq:multiobj_op}
\end{multline}

Usually, case $l=2k$ is considered. It means that the objectives remain without changes and form the space for multi-objective optimization. In this case $G=E$, where $E$ is the equality operator. The section below contains the particular realization used to obtain the PDE system using a combined evolutionary algorithm and sparse regression.

\section{Algorithm description}
\label{sec:algorithm_description}

\begin{figure*}[ht]
    \centering
    \includegraphics[width=0.7\textwidth]{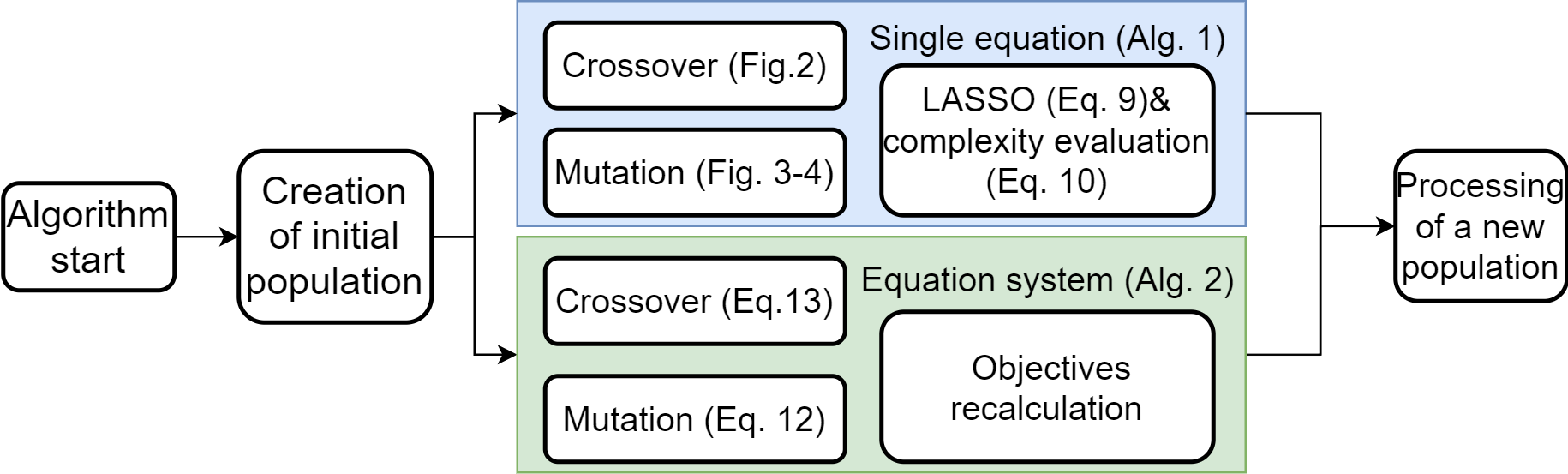}
    \caption{Principal scheme of the algorithm. Colors are representing the two different parts of the co-evolutionary algorithm.}
    \label{fig:main_scheme}
\end{figure*}

The developed algorithm of differential equation/system of differential equation discovery is composed of two elements: the tool for constructing a single equation (system of equations) according to the specific parameters and data, and the multi-objective optimization procedure, that detects the Pareto frontier of trade-offs between the previously introduced "quality" of the process representation and the "complexity" metrics of individual equations. The resulting algorithm is co-evolutionary and consists of two parts shown in the general scheme Fig.~\ref{fig:main_scheme}.

The single equation discovery part is a single-objective evolutionary algorithm that is described in detail in Sec.~\ref{sec:single_equation_discovery}, and the system discovery part is multi-objective and described in Sec.~\ref{sec:system_discovery}. Both parts are used simultaneously and form co-evolution.

\subsection{Evolutionary algorithm of equation discovery}
\label{sec:single_equation_discovery}

The construction of every equation begins with the definition of a set of function, i.e. tokens  $T = \{T_1, T_2, \; ... \; , T_n\}$, where each subset $T_j$ represents a separate class of function with specific properties, for example, derivatives of a specific components of velocity vector across all axes: $T_k = \{u, \frac{\partial u}{\partial x}, \frac{\partial u}{\partial y}, \frac{\partial u}{\partial z}, \frac{\partial u}{\partial t}, \frac{\partial^2 u}{\partial x^2}, \; ... \}$. The tokens will be used as the differential operator elements (most commonly - factors in the term) as in Eq.~\ref{eq:token_struct_equation}. The evolutionary algorithm of a single equation discovery, performs the construction of the data-driven equation in regards to minimization of discrepancy: $L_k \longrightarrow 0$. The pseudo-code for the procedure of system of equations discovery is presented on the Algorithm \ref{alg:ESTAR}.

\begin{equation}
    L_k =\sum_{i = 1}^{n\_terms} \alpha_i \prod_{j = 1}^{n\_factors} t_{ij}, \; t_{ij} \in T
\label{eq:token_struct_equation}
\end{equation}

\begin{figure}[h!]
    \centering
    \includegraphics[width=0.45\textwidth]{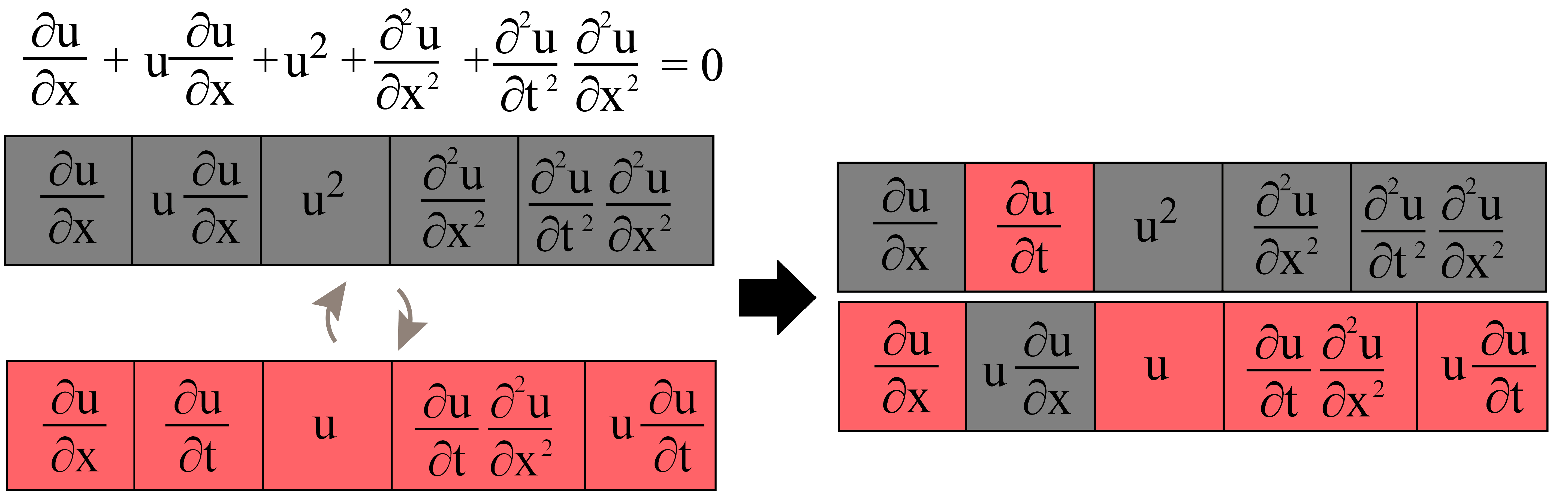}
    \caption{Individual form and recombination scheme of the single equation part of the algorithm. Grey and red colors are representing two different individuals.}
    \label{fig:single_xover}
\end{figure}

The encoding of the evolutionary algorithm individual is done in the following manner: a chromosome represents a candidate equation, where its elements, genes, representing factors of the equation, are divided into groups of random size (up to a specific limit) to represent the terms. Each factor can have alterable parameters that are also optimized during the equation construction (for example, a sine function can be used as a token, and it contains parameters: frequency, axis, along with that the sine is taken, and power). During the initialization of the algorithm, a set of candidate equations is randomly created. The example of the single equation is shown in Fig.~\ref{fig:single_xover}.

The weights $\alpha_i$ represent the candidate equation's coefficients and shall be optimized by additional technique. The weight vector's desired property is its sparsity, and it can be achieved with the sparse regression operator. When the equation structure is obtained from the evolutionary operator's step, each of the terms is approximated with the other terms taken as features by the LASSO regression, and the term with the best approximation is selected as the final "right side" of the equation. The non-zero coefficients with the corresponding terms compose the desired equation. This approach is introduced to exclude the unavoidable trivial structures with zero weights vectors, appearing with zero-valued right part approximation. Also, it is used to view all possible equations containing the same structure. However, LASSO regression can only obtain intermediate values of the equation coefficients $\beta$ due to the necessity to have data-centered and normalized. So, we have to solve the problems of minimizing the function Eq.~\ref{eq:Lasso} in respect to the coefficients $\beta$ for each equation term, selected as ``target''. Here, in the problem of approximation of $k$-th term, we have to evaluate the terms of the equation in points with known values to form the matrix $\mathbf{F}_k$ and vector of target term values ${F}_{target,k}$. 

\begin{equation}
\label{eq:Lasso}
\Vert{\mathbf{F}_{k}}\beta -{{F}_{target,k}}\Vert_{2}^{2}+\lambda {\Vert{\beta }\Vert_{1}} \longrightarrow    \min \limits_\beta
\end{equation}

After calculating coefficients $\beta$, we need to initiate the linear regression on original data to discover the correct values of the coefficients $\alpha$.

The fitness function is the inverse value of a norm of the best equation term approximation error as in Eq.~\ref{eq:fitness_val}. By maximizing that error, we perform the differential operator's search, which is close to zero.     

\begin{equation}
    \begin{split}
    f_{fitness} &= (|| L_k ||_2)^{-1} \\ 
    &= (|| \sum_{i = 1, i \neq t\_idx}^{n\_terms} \alpha_i \prod_{j = 1}^{n\_factors} t_{ij}  - \\ &- \prod_{j = 1}^{n\_factors} t_{t\_idx \; j} ||_2)^{-1}
    \end{split}
    \label{eq:fitness_val}
\end{equation}  

Both recombination and mutation are used as variation operators to improve the quality of the population. The recombination operator (Fig.~\ref{fig:single_xover}) operated differently on genes of two offspring, depending on the similarities of the parents' ones.

All terms are separated into three groups: the terms with the same tokens with the same parameters, the terms with the same tokens, the different parameters, and terms with unique tokens. The first group is not affected by crossover, the offspring duplicates of the terms from the second group have values of parameters as a proportion between their parent ones, and the terms of the third group can be swapped with defined probability.

\begin{figure}[h!]
    \centering
    \includegraphics[width=0.45\textwidth]{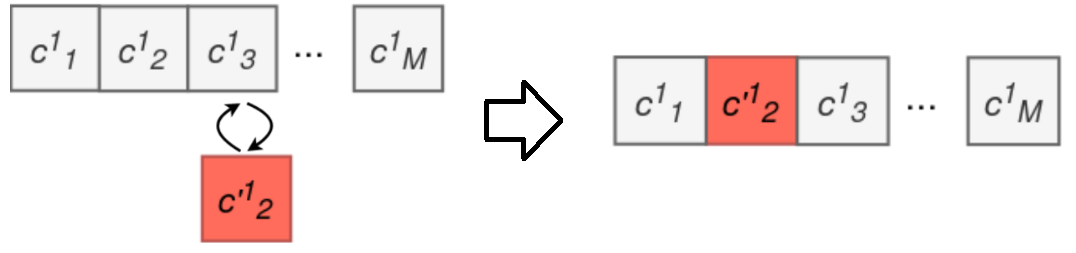}
    \caption{Mutation operator of the first kind. In this case whole term in the equation is replaced with another.}
    \label{fig:single_mutation_a}
\end{figure}

The mutation operator is implemented in two types. First, the gene can be changed to a new one, resulting in the change of an equation term or a factor inside a term as shown in Fig.~\ref{fig:single_mutation_a}. 

\begin{figure}[h!]
    \centering
    \includegraphics[width=0.45\textwidth]{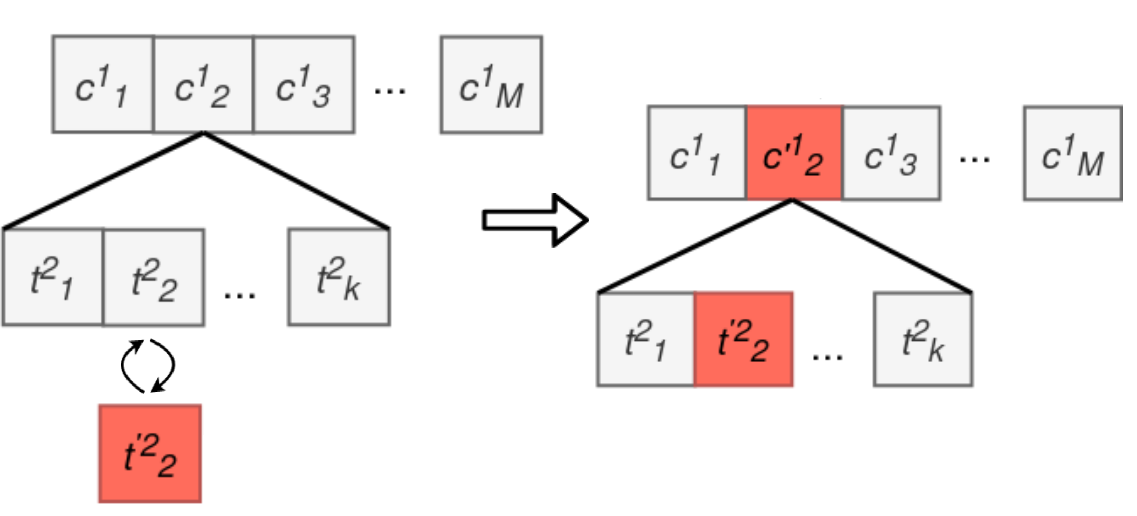}
    \caption{Mutation operator of the second kind. In this case term remains the same, however, either the parameters or differential order and variable may be altered or additional differential term may be inserted in the product.}
    \label{fig:single_mutation_b}
\end{figure}

Next, the mutation is used for the optimization of token parameters: if a token has alterable parameters (i.e., frequency of a sine function), a random increment from the normal distribution $\mathcal{N}(0, \sigma)$, with $\sigma$ as a fraction of allowed range for parameter, can be added as shown in Fig.~\ref{fig:single_mutation_b}. 

Several challenges appear with the shift from discovering a single equation to the derivation of differential equations systems. The construction of the system of equations is done in sequential order by equations. To avoid isolated equations, where each of the equations describes a separate variable with a probably overtrained structure, we try to connect the system's equations by introducing the variable change held by the resulting equation. After obtaining the value of "equation error" $L_i$, it is subtracted from the values of the new tokens as in Eq.~\ref{eq:variable_change}. 

\begin{equation}
    t_{ij} |_{k+1} = t_{ij} |_k - L_k = t_{ij} - \sum_{n=1}^{k}L_n
    \label{eq:variable_change}
\end{equation}

Also, we have introduced restrictions on the duplicated appearance of equations, describing the same variable, and only it (for example, a pair of equations $\frac{\partial f}{\partial x} + \frac{\partial f}{\partial y} = 0$, and $\frac{\partial f}{\partial t} = 0$), to force the equations to connect the variables. By penalizing the fitness value ($f_{fitness} = 0$, if it repeatedly only describes the same variable, as any of the previous equations) of such equations, we force the evolutionary to promote equations that describe all variables of a dynamic system.

\begin{algorithm}[ht!]
\label{alg:ESTAR}
 \KwData{set of tokens $T = \{T_1, T_2, \; ...\; T_n\}$, where subsets are divided into dependent and independent variables, sparsity constant for each equation $(\lambda_1, \; ... \;, \lambda_{n\_eq})$}
 \KwResult{system of differential equations} 
 \For{$equation\_idx = 1$ to $n\_var_{dependent}$}{
  Randomly generate initial population of candidate equations from tokens from set $T$\;
  \For{individual in population}{
    $right\_part\_idx = 0$\;
    $max\_fitness\_val = - inf$\;
    \For{target\_idx = 0 to terms\_number}{
     Apply sparse (LASSO) regression to find intermediate coefficients $\beta$\;
     Apply linear regression to find correct coefficients of the equations \& get fitness value $fit_val$\;
     \eIf{$fit\_val > max\_fitness\_val$}{
     $max\_fitness\_val = fit\_val$\;
     $right\_part\_idx = target\_idx$\;
     }{pass}
    }    
  }
  \For{epoch = 1 to epoch\_number}{
   population.sort()\;
   Remove worst solutions to maintain population size\;
   Tournament selection of parents for recombination\;
   Apply recombination and mutation operators\;
   \For{new\_individual in offsprings}{
    Use LASSO operator and linear regression to find best partition into left \& right parts and calculate fitness (as above)\;
   }
   Add new solutions into population\;
  }
  Replace variables: $t' = t - Lu, \; t \in T$
}

 \caption{The pseudo-code of system of equations discovery}
\end{algorithm}

\subsection{Obtaining the Pareto frontier of the equation systems} 
\label{sec:system_discovery}

In the previous developments, we have examined that by altering the algorithms' hyperparameters, mainly the LASSO regression's sparsity constant, utilized in the discovery of terms, we can shift the trade-off between quality and complexity. The problem of multi-objective optimization in space, created by the equations' quality and complexity metric, is solved by the Many-Objective Optimization Evolutionary Algorithm Based on Dominance and Decomposition (MOEA/DD), proposed in \cite{b8}.  

In other words, the multi-objective optimization algorithm operates as the meta-optimization for the main algorithm of system discovery. In this evolutionary process, the individuals represent systems of differential equations. The encoding operates as follows: the sparsity constants for each equation are combined into a vector $(\lambda_1, \lambda_2, ..., \lambda_{n\_eq},\; \lambda_i \in R^{+})$ (due to the properties of LASSO operator, sparsity constants are positive real values), which is considered an evolutionary algorithm chromosome and is affected by evolutionary operators during the search process. This representation's motivation is based on the notion that the equation discovery algorithm converges to the single solution with the specified hyperparameters. Therefore, with enough epochs for equation discovery, a vector of sparsity constants unequivocally defines a system of differential equations. 

In the initial stage of the evolution, according to the standard workflow of the MOEA/DD algorithm, we have to evaluate the best possible value for each of the objective functions: for complexity metric, it is reasonable to set the value to 0, while for the process representation quality (L2 norm of the vector of error in the grid points) the same assumption can be made only to a certain degree: the possible stochastic nature of processes or noise, present in measurements, limit the resulting quality. Therefore, a testing run of an equation discovery algorithm can be held to approximate the best possible quality of a solution. Next, to start the evolutionary search, we generate the population of solutions by finding systems with randomized sparsity constants and divide the search space into sections by weights vectors in the manner proposed in \cite{b9}. With the weights mechanism, the algorithm can preserve diversity in the population and cover the Pareto frontier with candidate points.

We use the conventional variation techniques with the introduced representation of individuals: mutation and recombination (crossover) operators. The mutation operator involves changing a candidate from the population with the addition of an increment from a normal distribution $\mathcal{N}(0, \sigma)$ to the specific sparsity constant in its gene with a pre-defined probability $p_{mut} \in (0, 1)$, as in the Eq.~\ref{eq:mutation_sys}.

\begin{equation}
    \begin{array}{cc}
     (\lambda_1, \lambda_2, \; ... \;, \lambda_{n\_eq}) \longrightarrow (\lambda'_1, \lambda'_2, \; ... \;, \lambda'_{n\_eq})  \\
      p_i \sim U(0,1) \\
         \text{if} \; p_i < p_{mut} \;\; \text{then} \; \lambda'_i = \lambda_i + \delta, \;  \delta \sim \mathcal{N}(0, \sigma) \\
         \text{else} \; \lambda'_i = \lambda_i
    \end{array}
    \label{eq:mutation_sys}
\end{equation}

The recombination operator involves the creation of new individuals using the selected parents. The offsprings should have characteristics resembling both their parents, which is implemented by the selection of offsprings genes' values in the diapason between their parents: the new values for each of the gene in the offsprings' chromosomes are selected as a weighted sum of their parents' ones, having coefficient $\alpha \in U(0, 1)$. The systems' recombination scheme is shown in Eq.~\ref{eq:xover_sys}.  

\begin{equation}
    \begin{array}{cccl}
           (\lambda^{1}_1, \lambda^{1}_2, \; ... \;, \lambda^{1}_{n\_eq}) \longrightarrow (\lambda'^{1}_1, \lambda'^{1}_2, \; ... \;, \lambda'^{1}_{n\_eq})  \\
         (\lambda^{2}_1, \lambda^{2}_2, \; ... \;, \lambda^{2}_{n\_eq}) \longrightarrow (\lambda'^{2}_1, \lambda'^{2}_2, \; ... \;, \lambda'^{2}_{n\_eq}) \\
         p_i \sim U(0,1)\\
         \text{if} \; p_i < p_{xover} \, \text{then} \,
         \lambda'^{1}_i = \alpha*\lambda^{1}_i + (1 - \alpha) * \lambda^{2}_i\\
         \text{else} \, \lambda'^{1}_i = \lambda^{1}_i, \;  \lambda'^{2}_i = \lambda^{2}_i
    \end{array}
    \label{eq:xover_sys}
\end{equation}

The selection of parents for the crossover is held for each objective function space region, defined by weights vectors. With a specified probability of maintaining the parents' selection, we can select an individual outside the processed subregion to partake in the recombination. In other cases, if there are candidate solutions in the region associated with the weights vector, we make a selection among them. The final element of MOEA/DD is population update after creating new solutions, which is held without significant modifications. 

\begin{algorithm}[h!]
\label{alg:MOEADD}
 \KwData{set of tokens $T = \{T_1, T_2, \; ...\; T_n\}$; subsets are divided into dependent and independent variables; objective functions: quality and complexity of systems}
 \KwResult{Pareto frontier, composed of systems of differential equations}
 Create a set of weight vectors $\mathbf{w} = (w^1, ..., w^{n\_weights}), \; w^i=(w^i_1, \; ..., \; w^i_{n\_eq + 1}) $\;
 \For{weight\_vector in weights}{
 Select K nearest weight vectors to the weight\_vector\;
 }
 Randomly generate a set of candidate systems of equations (using Alg.~\ref{alg:ESTAR}) \& divide them into non-dominated levels\;
 Divide the initial population into groups by subregion, to which they belong\;
 \For{epoch = 1 to epoch\_number}{
  \For{weight\_vector in weights}{
   Parent selection\;
   Apply recombination to parents pool and mutation to individuals inside the region of weights\;
   \For{offspring in new\_solutions}{
    Optimize the structure of systems of equations for offspring\;
    Get values of objective functions for offspring\;
   }
   Update population\;
  }
 }

 \caption{The pseudo-code of Pareto frontier construction, using adapted MOEA/DD algorithm}
\end{algorithm}

Both algorithms Alg.~\ref{alg:ESTAR} and Alg.~\ref{alg:MOEADD} are used simultaneously in a co-evolution manner to obtain the Pareto frontier of the systems based on observational input data.

\section{Experimental studies}
\label{sec:experiment}

To analyze the algorithm's performance, we have applied it to the synthetic data set, representing a solution to the known system of partial differential equations. For that task, we have selected a Navier-Stokes system of equations with assumptions of an incompressible fluid, displayed on Eq.~\ref{eq:Navier_Stokes}, that describes relations between velocity (with velocity components $u, \; v$), and pressure $p$, and describes a flow of liquid in a pipe. Here, $\Delta = \frac{\partial^2}{\partial x^2} + \frac{\partial^2}{\partial y^2}$ is the del operator, and $\nabla = \frac{\partial}{\partial x} + \frac{\partial}{\partial y}$ is the Laplace operator, $\rho$ is density, that is assumed to be constant in the domain, and $\mathbf{F} = (\vec{F}_x, \vec{F}_y)$ is the mass force.

\begin{equation}
    \left\{\begin{array}{cc}
         \frac{\partial u}{\partial t}=\frac{1}{\rho} \nu \nabla^2 u + \vec{F}_x - \frac{1}{\rho} \frac{\partial p}{\partial x}\\
         \frac{\partial v}{\partial t}=\frac{1}{\rho} \nu \nabla^2 v + \vec{F}_y - \frac{1}{\rho} \frac{\partial p}{\partial y} \\
         \Delta p = g
    \end{array}
    \right. 
\label{eq:Navier_Stokes}
\end{equation}

The system of equations was solved on the domain of 0 to 100 with 100 grid nodes at each spatial domain, and from 0 to 10 with 400 grid points in time. The initial conditions for the velocity are set in Eq.~\ref{u_inital} \& Eq.~\ref{v_inital}, while boundary conditions on area limits $\Gamma$ were set as $u|_{\Gamma} = 0$; $v|_{\Gamma} = 0$. Other parameters have the following values: $F_x = F_y = 0.01$; density $\rho = \frac{1}{10}$; viscosity $\nu = 1.0$. Boundary condition for pressure was set as $p|_{y = 0; \;y = 100}=4.9*x^2 - 490x$; $p|_{x = 0; \;x = 100} = 0$.

\begin{multline}
    u |_{t=0} = 10^3 \sin{(\frac{1}{10^4}  x (100-x))} \\ \sin{(\frac{1}{10^4} y (100-y))} \sin{(\frac{pi}{25} x)}
    \label{u_inital}
\end{multline}
\begin{multline}
    v |_{t=0} = 10^3 \sin{(\frac{1}{10^4}  x (100-x))} \\ \sin{(\frac{1}{10^4} y (100-y))} \sin{(\frac{\pi}{25} y)}
    \label{v_inital}
\end{multline}

From the numerical solution to the system (matrices of velocity components and pressure values, calculated for a grid in the studied domain), the algorithm's input data, represented by values of derivatives, evaluated on the grid's points obtained by numerical differentiation. Earlier works have shown that the least errors in the derivatives' values are achieved in the experiments, where the differentiation is done with the analytical differentiation of polynomials, fit to the data. Thus, the building blocks of the equations are represented by functions $u, v, p$, and their derivatives across two spatial axes and time up to 3-rd order $\frac{\partial u}{\partial x}, \frac{\partial u}{\partial y}, \frac{\partial u}{\partial t}, \frac{\partial v}{\partial x}, \; ... \;, \frac{\partial^3 p}{\partial t^3}$. 

However, with the blind inclusion of all derivatives, we can face issues with the discovery of simplified equations in the forms of $\frac{\partial f}{\partial x_i} = 0$, when the function $f$ does not depend on the variable $x_i$. For example, in our case, the pressure field is static. Thus some equations like Eq.~\ref{eq:static_pressure} with arbitrary order of derivative $n$ can be discovered with very high fitness values.

\begin{equation}
    \frac{\partial^n p}{\partial t^n} = 0
    \label{eq:static_pressure}
\end{equation}

\begin{figure}[ht!]
\includegraphics[width=0.49\textwidth]{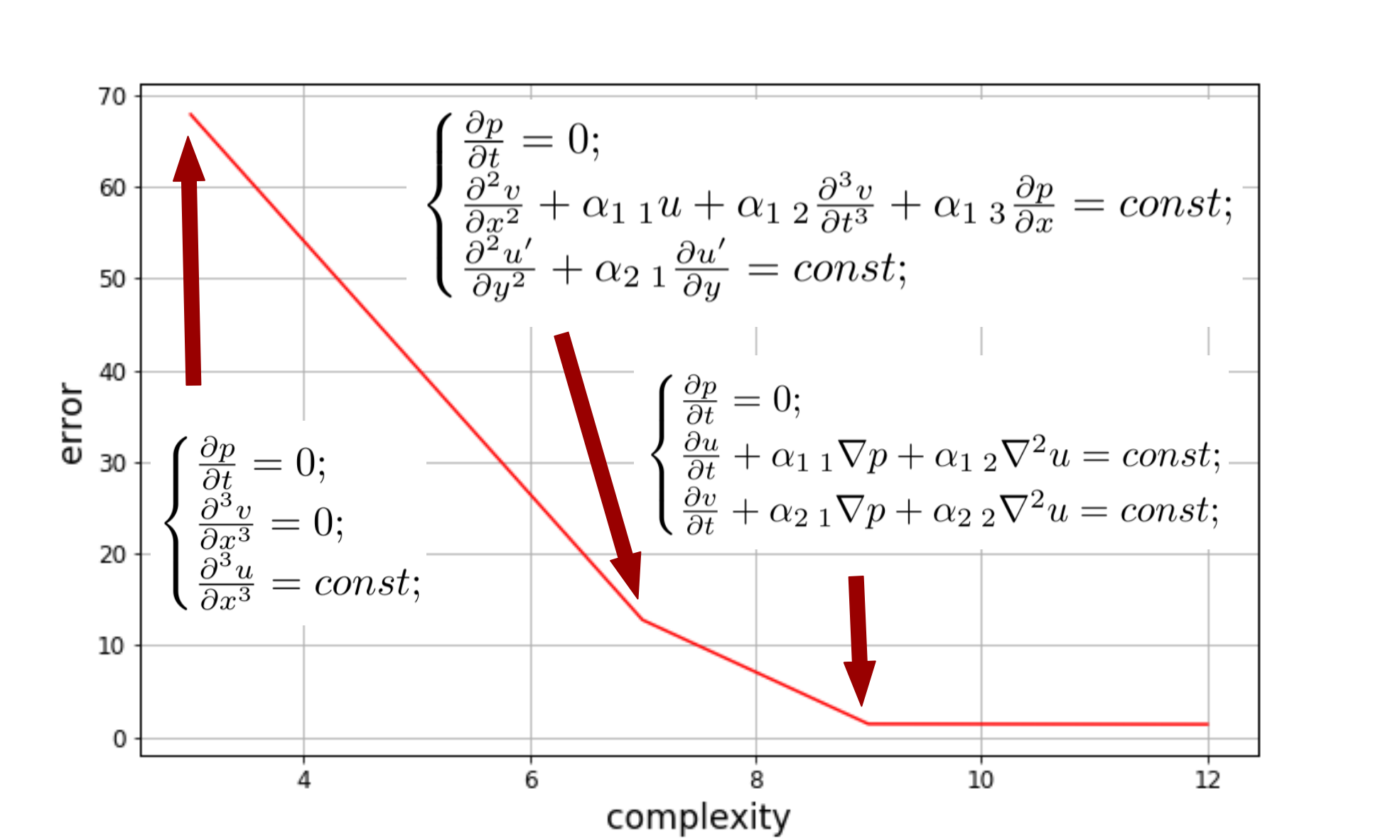}
\caption{The Pareto frontier of quality (total L2-norm of the equations)-complexity (total number of terms in the equations) trade-offs for systems of equations}
\label{fig:pareto_equations}
\end{figure}

The analysis of the obtained Pareto frontier and the comparison between discovered equations indicate that the experiment data can be described with multiple equations with different quality degrees. In Fig.~\ref{fig:pareto_equations}, the complexity metric shows a number of terms except for constant across all equations of the discovered system, while the error indicates the sum of L2-norms of a discrepancy of the corresponding equation of the systems. 

Before analyzing the Pareto frontier, we note that when the equations contain only one term except constant, the most straightforward case provides the system description's highest error. More complex systems describe the dynamics with higher quality, while after the construction of momentum conservation law from the Navier-Stokes system of equation, coupled with an equation for pressure, the addition of new terms does not reduce the model's error. However, the correct equation for pressure was not constructed due to the equation's good quality containing the time derivative of pressure. 

The three cases in Fig.~\ref{fig:pareto_equations} show the possibility of tuning the resulting system. With the modern regression methods, we may obtain the Navier-Stokes vector equation in the form Eq.~\ref{eq:vector_Navier_Stokes}.

\begin{equation}
         \frac{\partial \vec{u}}{\partial t}-\frac{1}{\rho} \nu \vec{\nabla}^2 \vec{u}=const
\label{eq:vector_Navier_Stokes}
\end{equation}

We can obtain the additional pressure equation. However, the expert is required to show that the third pressure equation exists. 

We note that the proposed algorithm works autonomously. First system highlighted in Pareto frontier has the form Eq.~\ref{eq:res_pareto_1}.

\begin{equation} 
\begin{cases}
    \frac{\partial p}{\partial t} = 0; \\
    \frac{\partial^3 v}{\partial x^3} = 0; \\
    \frac{\partial^3 u}{\partial x^3} = const;
\end{cases}
\label{eq:res_pareto_1}
\end{equation}

System Eq.~\ref{eq:res_pareto_1} shows the case described above. Simple trivial equations have the lowest possible complexity and good quality since the analytical differentiation methods result in zero fields. However, since the numerical differentiation is used, non-zero fields are obtained, and overall non-zero quality results from the differentiation error. It cannot be filtered out automatically since it is unknown whether the system is simple or the numerical differentiation scheme introduces the error.

Second highlighted system has the form Eq.~\ref{eq:res_pareto_2}.

\begin{equation} 
\begin{cases}
    \frac{\partial p}{\partial t} = 0; \\
    \frac{\partial^2 v}{\partial x^2} + \alpha_{1 \; 1} u + \alpha_{1 \; 2} \frac{\partial^3 v}{\partial t^3} + \alpha_{1 \; 3} \frac{\partial p}{\partial x} = const; \\
    \frac{\partial^2 u}{\partial y^2} + \alpha_{2 \; 1} \frac{\partial u}{\partial y}  = const;
\end{cases}
\label{eq:res_pareto_2}
\end{equation}

At the point on the Pareto frontier that gives the system Eq.~\ref{eq:res_pareto_2}, the separate pressure equation is discovered. The two other equations represent the data-specific equations, which give higher quality than the Navier-Stokes system. However, their form is very sensitive to the changes in data such as noise, and this system may not be considered as the stable one.

The third highlighted system has the form Eq.~\ref{eq:res_pareto_3} which is similar to the Navier-Stokes system Eq.~\ref{eq:Navier_Stokes}.

\begin{equation} 
\begin{cases}
    \frac{\partial p}{\partial t} = 0; \\
    \frac{\partial u}{\partial t} + \alpha_{1 \; 1} \nabla p + \alpha_{1 \; 2} \nabla^2 u = const;\\
    \frac{\partial v}{\partial t} + \alpha_{2 \; 1} \nabla p + \alpha_{2 \; 2} \nabla^2 u = const; \\
\end{cases}
\label{eq:res_pareto_3}
\end{equation}

We achieve the automatic pressure equation separation together with the complete set of component-wise equations. As seen, the analysis of the Pareto frontier makes the interpretation of the system discovery process clear. We may change the optimization in a more precise manner to obtain the system, representing the general data-driving laws such as the Navier-Stokes equations.

\section{Conclusion}
\label{sec:conclusion}

In this paper, we have proposed an algorithm of differential equations systems discovery. We combine the evolutionary algorithm of obtaining a system with the multi-objective optimization problem to provide better insight into a discovered equation's ability to model the studied process. While the introduced algorithm of equation or system of equations derivation can get the best structure possible with the defined input tokens. In ideal conditions, we will determine which resulting equation we would like to use for further simulation during the physical process modeling. 

The algorithm has the following properties:

\begin{itemize}
    \item The equations are obtained in a non-vector form that allows tuning the models more subtly. Moreover, it allows to control the process of the equation discovery;
    \item Introduced evolutionary operators allow to avoid the pre-defined library of differential terms that increases the variety of resulting equations;
    \item Multi-objective optimization allows to tune every equation in the system and obtain Pareto frontier of the models that increase the expert interpretation possibilities.
\end{itemize}

The further work on the related topic will be based mainly on the area of algorithm convergence to the complete structures that describe a process, avoiding overtraining as well as simplified solutions, such as the ones faced in the experiments, and the area of improving links between the equations of the system. Also, algorithm performance shall be addressed to make its application easier for the researcher with ordinary computational powers.

\section{Code and Data availability}
\label{sec:repo}
The numerical solution data and the Python code that partially reproduce the experiments are available at the GitHub repository \footnote{https://github.com/ITMO-NSS-team/FEDOT.Algs/blob/master/estar/}.

\end{document}